\documentclass{article}
\pdfoutput=1
\usepackage[utf8]{inputenc} 
\usepackage[T1]{fontenc}    
\usepackage{hyperref}       
\usepackage{url}            
\usepackage{booktabs}       
\usepackage{amsfonts}       
\usepackage{nicefrac}       
\usepackage{microtype}      
\usepackage{amsmath}
\usepackage{multirow}
\usepackage{bm}
\usepackage{graphics}
\usepackage{graphicx}
\usepackage{xcolor}
\usepackage{authblk}
\usepackage{wrapfig}
\usepackage{lipsum}
\usepackage{setspace}

\usepackage[nonatbib,final]{emc2_2020}

\title{The Architectural Implications of Distributed Reinforcement Learning on CPU-GPU Systems}

\author[1]{\textbf{Ahmet Inci}}
\author[2]{\textbf{Evgeny Bolotin}}
\author[2]{\textbf{Yaosheng Fu}}
\author[2]{\textbf{Gal Dalal}}
\author[2]{\textbf{Shie Mannor}}
\author[2]{\\ \textbf{David Nellans}}
\author[1,3]{\textbf{Diana Marculescu}}
\affil[1]{Carnegie Mellon University}
\affil[2]{NVIDIA}
\affil[3]{The University of Texas at Austin}

\begin{document}

\maketitle

\begin{abstract}

With deep reinforcement learning (RL) methods achieving results that exceed human capabilities in games, robotics, and simulated environments, continued scaling of RL training is crucial to its deployment in solving complex real-world problems. However, improving the performance scalability and power efficiency of RL training through understanding the architectural implications of CPU-GPU systems remains an open problem. In this work we investigate and improve the performance and power efficiency of distributed RL training on CPU-GPU systems by approaching the problem not solely from the GPU microarchitecture perspective but following a holistic system-level analysis approach. We quantify the overall hardware utilization on a state-of-the-art distributed RL training framework and empirically identify the bottlenecks caused by GPU microarchitectural, algorithmic, and system-level design choices. We show that the GPU microarchitecture itself is well-balanced for state-of-the-art RL frameworks, but further investigation reveals that the number of actors running the environment interactions and the amount of hardware resources available to them are the primary performance and power efficiency limiters. To this end, we introduce a new system design metric, \textit{CPU/GPU ratio}, and show how to find the optimal balance between CPU and GPU resources when designing scalable and efficient CPU-GPU systems for RL training.
\end{abstract}

\vspace{-2mm}
\section{Introduction}
\vspace{-2mm}
Reinforcement learning (RL) algorithms have recently achieved remarkable success across a variety of tasks. With the recent improvements in CPU, GPU, and TPU architectures, deep reinforcement learning methods are able to exceed human performance in various complex games such as Go \cite{alphago}, Dota 2 \cite{dota}, and StarCraft II \cite{vinyals2017starcraft}. However, despite the impressive ability to learn complex policies, state-of-the-art RL algorithms require an increasingly large number of training iterations to successfully learn even simple games \cite{badia2020agent57}. Therefore, to accelerate scientific discovery via reinforcement learning on real-world problems it is imperative to shorten the experiment turnaround time via \textit{scalable} and \textit{efficient} RL training.

Despite recent efforts focusing on improving RL efficiency and throughput on specialized hardware accelerators such as TPUs \cite{reverb,seed,impala,acme}, scaling RL training performance and understanding the hardware implications of CPU-GPU designs is still an open research problem \cite{stooke2018accelerated}. 
In this work we investigate and optimize the performance and power efficiency of distributed RL training on CPU-GPU systems by performing an in-depth analysis at both the GPU microarchitectural and CPU-GPU system levels. 

This work makes the following contributions:
\begin{itemize}
    \item We quantify the detailed GPU hardware utilization for state-of-the-art RL algorithms and empirically identify the main performance bottlenecks occurring from system-level architectural limitations and/or algorithmic implementation choices. Our analysis shows that GPU microarchitecture itself is well-optimized for state-of-the-art RL training \cite{seed}. 
    \item We identify and demonstrate that environment interaction, and the number of CPU hardware resources available for this task, are the primary performance and power efficiency limiters for end-to-end RL training. This suggests that further research on system-level optimizations rather than GPU microarchitecture optimizations are necessary to achieve the best performing and the most power-efficient systems for RL. 
    \item We introduce and explore a new system design metric, \textit{CPU/GPU ratio}, that will aid in designing scalable and efficient future CPU-GPU systems for RL training. We show that performance and energy-efficiency of CPU-GPU systems can be improved by tuning the ratio of CPU hardware threads and GPU streaming multiprocessor (SM) in systems. 
\end{itemize}

\vspace{-2mm}
\section{Background and Related Work}
\begin{wrapfigure}{r}{0.4\textwidth}
    \centering
    \includegraphics[width=0.4\columnwidth]{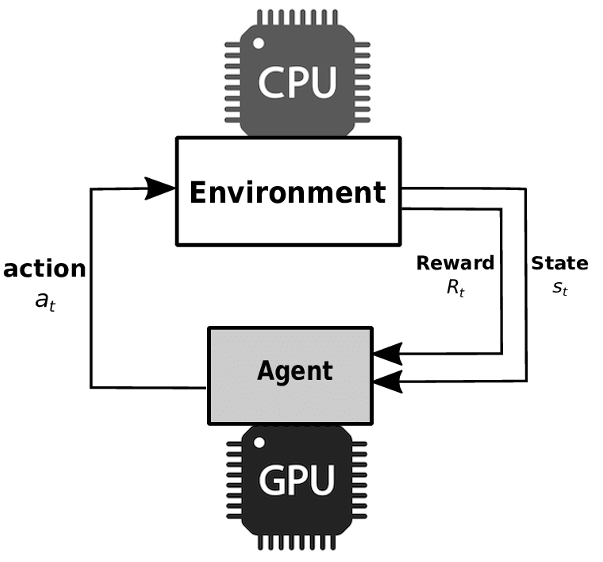}
    \caption{System-level mapping of reinforcement learning workloads on hardware platforms.}
    \label{fig:system}
    \vspace{-2mm}
\end{wrapfigure}

\vspace{-2mm}
Reinforcement learning algorithms largely consist of an agent interacting with an environment, with the objective of maximizing cumulative long-term reward (Figure \ref{fig:system}). In each discrete time step, the agent observes the current state $S_t$ and takes an action $a_t$. Based on $a_t$, the agent receives a reward $R_t$ and an observation that encodes a partial view of the environment's state. From a hardware execution standpoint, the environment interaction is traditionally handled by the CPU, whereas the agent that predicts the action to be performed on the environment is located on the GPU. 
Although there are examples of GPU-based environment interactions for physics simulations \cite{gpu_robotic_viktor}, the vast majority remain CPU-based.

\subsection{Distributed Reinforcement Learning}

With the advent of deep RL \cite{mnih2013playing}, the community has turned to distributed RL algorithms for processing the vast amounts of environment data need for training of deep RL models with acceptable turnaround times. RL algorithms are inherently heterogeneous and employ various parallel tasks such as environment interaction, model inference, model training, and replay buffer management \cite{reverb,replay_buffer}. Several research teams have been focusing on improving distributed deep RL training performance recently. For example, IMPALA \cite{impala} is a distributed RL training architecture which uses a novel off-policy correction algorithm (V-trace) to sequentially learn from the stream of environment interactions, generated by a large number of independent actors. 
Recently, SEED RL \cite{seed} achieved state-of-the-art performance on Arcade Learning Environment (ALE) \cite{Bellemare_2013} by proposing a central inference method which moves inference batches to the learner from the actor and uses multiple environments on the actor. This proved to be effective using R2D2 \cite{kapturowski2018recurrent}, a SOTA Q-learning algorithm. 

\vspace{-2mm}
\section{Methodology}
\vspace{-2mm}
We focus on the state-of-the-art, highly distributed SEED RL \cite{seed} framework and run experiments on the SEED RL implementation of R2D2 \cite{kapturowski2018recurrent} for the Arcade Learning Environment \cite{Bellemare_2013}. Our experimental platform is a NVIDIA DGX-1 station that is equipped with 8 V100 GPUs, each with 80 SMs and the system having a 20-Core Intel Xeon E5-2698 v4 CPU running at 2.2 GHz, providing 40 total hardware CPU threads. We use NVIDIA's \textit{nvprof} profiling tool to quantify the CPU-GPU utilization and dissect the hardware implications of RL-training on complex real-world machines and carry out GPU microarchitectural experiments using \textit{NVArchSim}, NVIDIA's highly accurate trace-driven architectural simulation infrastructure \cite{nvas}.
\vspace{-2mm}
\section{Experiments}
\vspace{-2mm}
In this work, we investigate the following three questions in the context of RL:  First, what is the actual GPU utilization for SOTA distributed RL training and what are the specific GPU microarchitectural bottlenecks? Second, if not the GPU itself, then what are the system-level performance and power efficiency bottlenecks? Finally, what should be the primary guidelines when architecting CPU-GPU systems for RL training? 

To this end, we first perform detailed hardware profiling and architectural simulations to identify if the GPU is the bottleneck for state-of-the-art distributed RL training and to better understand the GPU hardware utilization. Then, we perform a similar analysis for the CPU portion of the workloads to identify the primary performance bottlenecks in the CPU-GPU system. Finally, we conduct experiments to find the desired balance between CPU and GPU resources that optimize overall system throughput.

\paragraph{GPU Utilization And Performance Bottleneck Analysis on SOTA RL.}

To identify the performance bottlenecks of RL training on CPU-GPU systems, we first analyze NVIDIA's V100 GPU architecture via detailed hardware profiling and architectural simulations using NVIDIA's \textit{nvprof} profiling tool and \textit{NVArchSim} architectural simulator. 

\begin{wrapfigure}{r}{0.4\textwidth}
    \centering
    \includegraphics[width=0.4\columnwidth]{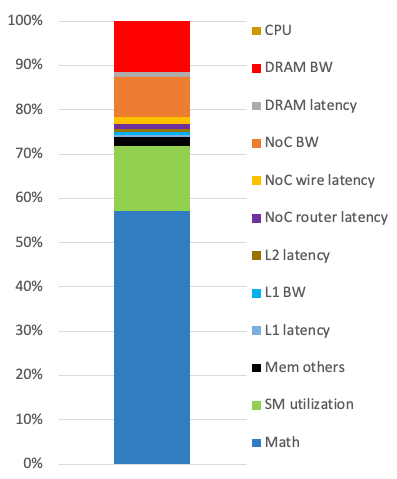}
    \caption{GPU hardware performance bottleneck breakdown for state-of-the-art SEED RL framework \cite{seed}.}
    \vspace{-2mm}
    \label{fig:hw_bottleneck}
\end{wrapfigure}

Figure \ref{fig:hw_bottleneck} presents a GPU hardware bottleneck analysis for the state-of-the-art SEED RL \cite{seed} framework on an NVIDIA V100 GPU. To identify the causes of performance inefficiency, we break down the overall execution time by GPU microarchitectural component, where each segment represents the performance overhead attributed to the component during execution. For example, the green bar shows the performance reduction caused by SM underutilization, when compared to a 100\% utilized SM. Similarly, the red bar represents the performance overhead introduced by a non-ideal DRAM bandwidth when compared to a GPU with infinite DRAM bandwidth. 
To assess the impact of each element on performance we calculate the relative overheads of each component by idealizing them sequentially, starting from the outermost external DRAM through the innermost SM component. 

Starting with the baseline GPU configuration, we first idealize the DRAM bandwidth to calculate the performance reduction caused by insufficient off-chip bandwidth. Then with the DRAM bandwidth unbounded, we set the DRAM latency to zero to measure the performance overhead due to data dependency stalls in DRAM. We repeat this until the entire memory system is idealized and all memory requests are returned immediately upon issue from the SM. Next, we simulate a GPU with a single SM and an idealized memory system to measure the effect of SM underutilization due to load imbalance or insufficient parallelism to keep all SMs occupied. We normalize this simulated throughput by the total number of SMs in the system to calculate the theoretical GPU performance as if all SMs were able to achieve full occupancy without load imbalance. 

Figure \ref{fig:hw_bottleneck} shows that Math (actual compute throughput), SM utilization, and DRAM bandwidth contribute 57\%, 15\%, and 12\% respectively to overall GPU execution time limitation.  This means that even with a zero latency, unbounded bandwidth, GPU with no SM utilization there is a less than a factor of two performance improvement possible through GPU microarchitectural optimization. We conclude that based on today's memory technologies the GPU microarchitecture itself is well-balanced for the SOTA RL framework. Therefore, we now shift our focus to CPU-GPU system-level analysis for RL training.

\textbf{Conclusion 1:} \textit{Architects should not solely focus on GPU microarchitecture and instead focus on system-level architectural parameters when optimizing systems for RL training.}

\paragraph{System-Level Performance And Power Efficiency.}

\begin{figure}[ht]
    \centering
    \includegraphics[width=0.48\textwidth]{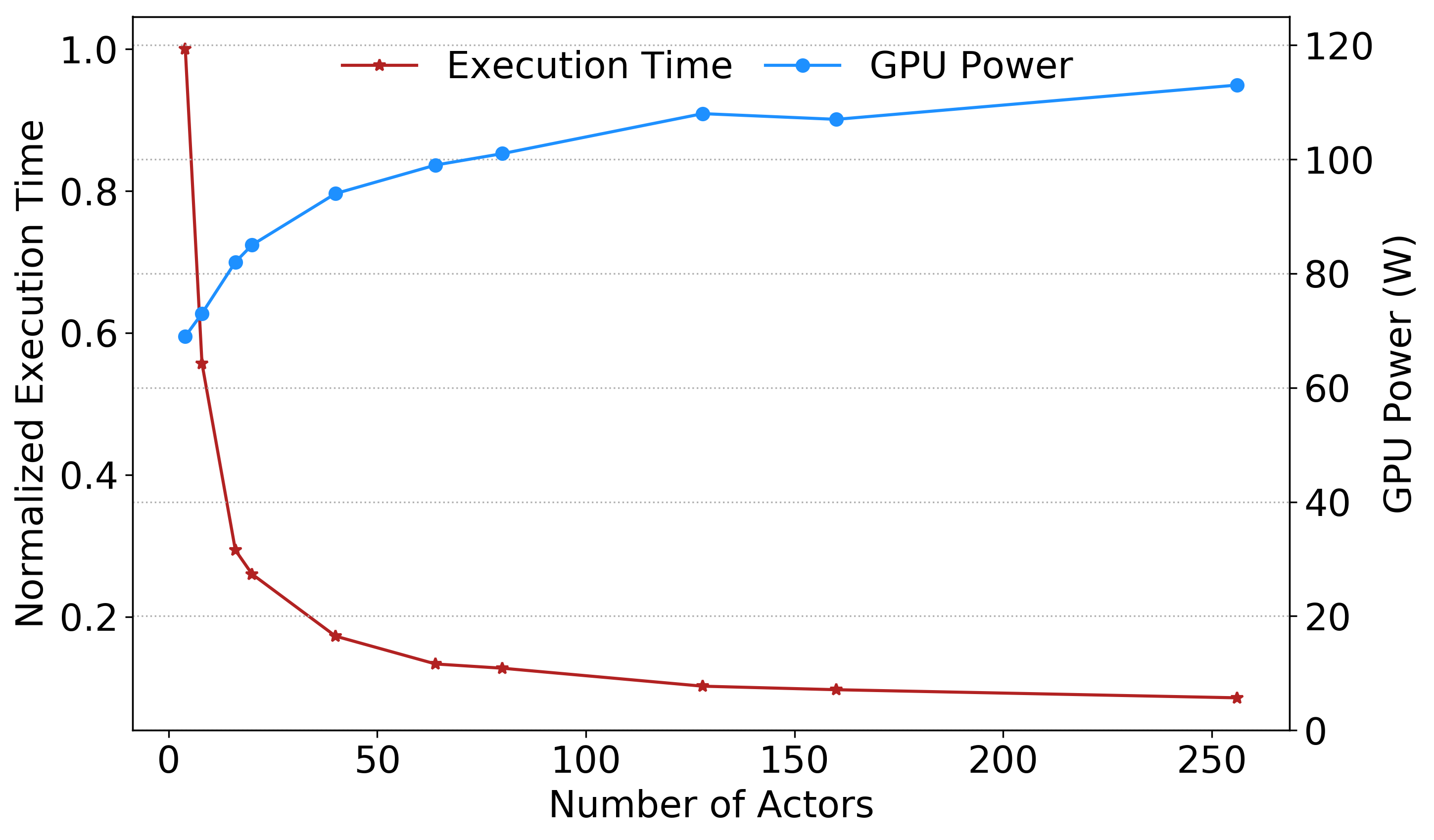}
    \includegraphics[width=0.49\textwidth]{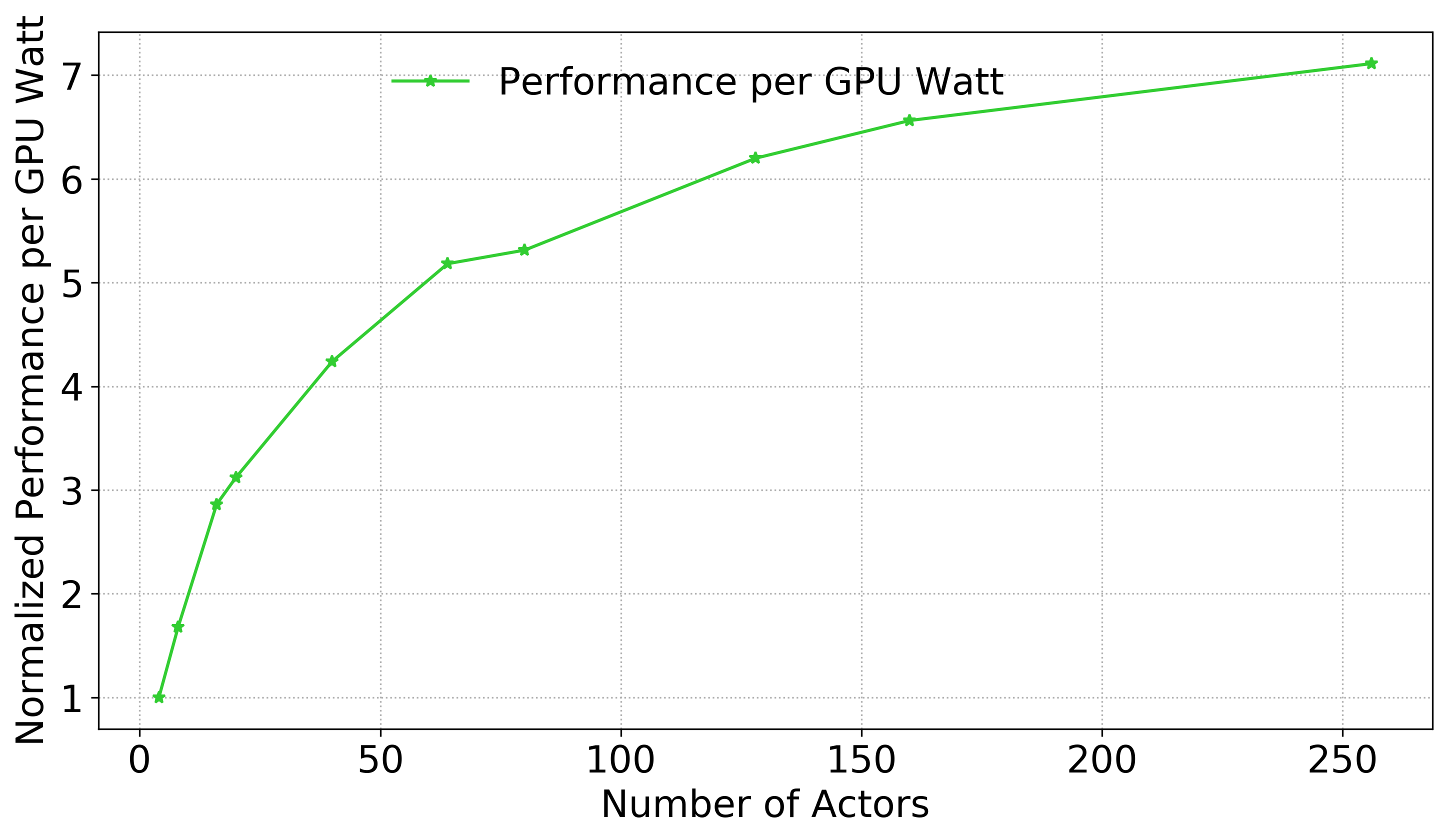}
    \caption{Impact of number of actors on runtime, GPU power (left), and Perf. per GPU Watt (right).} 
    \label{fig:actors}
    \vspace{-5mm}
\end{figure}

To understand the importance of CPU-GPU interactions and relative performance, we swept the number of simulation actors to investigate the impact on runtime and GPU power consumption on an NVIDIA V100 GPU. Figure \ref{fig:actors} shows the impact of the number of actors on normalized execution time and GPU power (left), and normalized performance per GPU Watt (right). Scaling the number of CPU-based actors from 4 to 40 provides a relative $5.8 \times$ speedup, whereas when scaling from 40 to 256 actors only $2 \times$ of additional speedup is achieved even though the number of actors is increased by $6.4 \times$. We conclude that performance scalability is primarily limited by the total number of CPU hardware threads, which is 40 in our system. Although scaling up the number of actors achieves impressive performance improvements for up to 40 actors, it yield diminishing returns beyond that point likely due to CPU resources being fully saturated.

Figure \ref{fig:actors} also shows that GPU power consumption (right y-axis) increases when growing the number of CPU-actors. This is because with an increased number of actors the system is able to finish each round of environment interactions faster, increasing GPU utilization. However, overall GPU power efficiency (performance per Watt) improves with the growing number of actors and overall energy per task decreases. 
This is because GPU power consumption at low utilization is relatively high ($\approx$70 W) and performance increases faster than the additional power consumption. This suggests that architectural to improve GPU power management and efficiency at low utilization could improve overall system efficiency.

\textbf{Conclusion 2:} \textit{System bottlenecks that limit environment interaction are the primary performance and power efficiency limiters in RL training. Efficiently scaling the number of actors and hardware available to execute them is critical to achieving optimal system performance and energy efficiency.}

\begin{wrapfigure}{r}{0.5\textwidth}
    \vspace{-0.5cm} 
    \centering
    \includegraphics[width=0.5\textwidth]{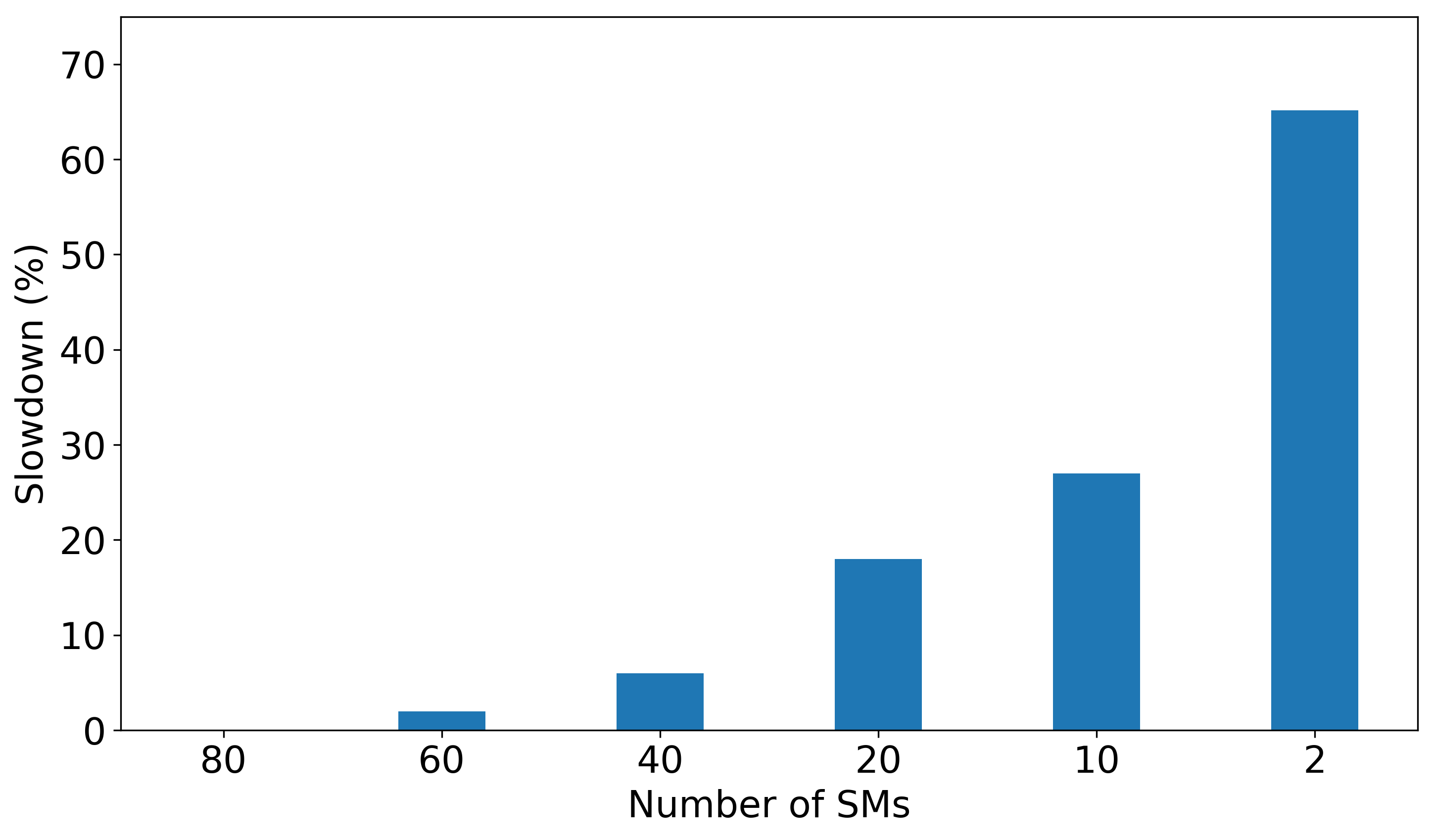}
    \caption{Performance slowdown when reducing the number of GPU SMs by limiting the number of SMs visible to the GPU-HW scheduler.}
   \label{fig:cpu_sm}
   \vspace{-0.4cm} 
\end{wrapfigure}

\vspace{-2mm}
\paragraph{Designing Scalable and Efficient CPU-GPU Systems for RL Training.}
To strike the right balance between CPU and GPU hardware resources for RL training, we propose a new system design metric, \textit{CPU/GPU ratio} -- which is the ratio between CPU hardware threads and total GPU SMs in the system.
While ideally we could add additional CPU resources to a system to understand the impact of this ratio, we are limited by the number of CPU hardware threads in our system which is hard to modify.  
Instead, we explore this by gradually disable SMs on our DGX-1 V100 GPU to mimic further scaling of \textit{CPU/GPU ratio} imitating a system with a larger number of CPU cores. 

Figure \ref{fig:cpu_sm} shows these measured slowdown with respect to a baseline NVIDIA V100 configuration with 80 SMs, a \textit{CPU/GPU ratio} of 1/2. Running with half the number of SMs (40 SMs, \textit{CPU/GPU ratio} = 1) yields only 6\% slowdown which implies large GPU resource underutilization due to insufficient actor throughput at the system-level, though if the number of SMs is reduced too much the GPU can become the bottleneck at the system-level (e.g. 2 SMs).
Existing CPU-GPU systems, such as the DGX-1 contain 8 NVIDIA V100 GPUs, each with 80 SMs,  yet only 20 CPU cores (40 hardware threads) resulting in \textit{CPU/GPU ratio} of 1/16. 

Recently, the DGX-A100 has improved \textit{CPU/GPU ratio} to 1/4 which is still insufficient as our results show that the \textit{CPU/GPU ratio} needs to be at least one (or higher) to achieve good efficiency. We conclude that future well-balanced RL systems will require a $16 \times$ and $4 \times$ improvement in \textit{CPU/GPU ratio} (compared to DGX-1 and DGX-A100 systems) respectively to maximize performance and power efficiency.

\textbf{Conclusion 3:} \textit{CPU/GPU ratio is a convenient rule of thumb to help system architects to design scalable and efficient CPU-GPU based systems for RL. To maximize actor throughput, RL systems should be equipped a number of CPU hardware threads equal to or higher than the number of GPU SMs in the system (for current generation SMs). This leads to the highest GPU utilization and the ultimately the lowest energy consumption per training task for the system.}
\vspace{-2mm}
\section{Conclusion}
\vspace{-2mm}
In this work we investigate and identify the primary system-level design parameters for improving the performance and power efficiency of distributed RL training on CPU-GPU based platforms. We approach the task not just from the GPU microarchitecture perspective but from a system-level perspective. Our findings indicate that the current GPU microarchitectures are already well-balanced for the state-of-the-art distributed RL. We show that environment interactions and the hardware resources dedicated to it are the primary performance and power efficiency bottlenecks in today's systems. We propose a new design metric, \textit{CPU/GPU ratio}, for balancing system CPU and GPU resources to design scalable and power efficient platforms for RL training. We conclude that these systems should be equipped with CPU hardware thread resources that are equal to or higher than the number of GPU SMs, thus enabling significant improvements in performance and power efficiency for distributed RL training workloads compared to many of today's systems.

\vspace{-2mm}

\bibliographystyle{plain}
\small{
\bibliography{main}

\begin{thebibliography}{10}

\bibitem{badia2020agent57}
Adria~Puigdomenech Badia, Bilal Piot, Steven Kapturowski, and et~al.
\newblock {Agent57}: Outperforming the atari human benchmark.
\newblock {\em arXiv preprint arXiv:2003.13350}, 2020.

\bibitem{Bellemare_2013}
Marc~G. Bellemare, Yavar Naddaf, Joel Veness, and et~al.
\newblock {The} arcade learning environment: An evaluation platform for general
  agents.
\newblock {\em Journal of Artificial Intelligence Research (JAIR)},
  47:253–279, 2013.

\bibitem{reverb}
Albin Cassirer, Gabriel Barth-Maron, Thibault Sottiaux, and et~al.
\newblock {Reverb}: An efficient data storage and transport system for ml
  research.
\newblock \url{https://github.com/deepmind/reverb}, 2020.

\bibitem{seed}
Lasse Espeholt, Raphael Marinier, Piotr Stanczyk, and et~al.
\newblock {S}eed rl: Scalable and efficient deep-rl with accelerated central
  inference.
\newblock In {\em International Conference on Learning Representations (ICLR)},
  2020.

\bibitem{impala}
Lasse Espeholt, Hubert Soyer, Remi Munos, and et~al.
\newblock {I}mpala: Scalable distributed deep-rl with importance weighted
  actor-learner architectures.
\newblock In {\em International Conference on Machine Learning (ICML)}, 2018.

\bibitem{acme}
Matt Hoffman, Bobak Shahriari, John Aslanides, and et~al.
\newblock {Acme:} a research framework for distributed reinforcement learning.
\newblock {\em arXiv preprint arXiv:2006.00979}, 2020.

\bibitem{replay_buffer}
Long ji~Lin.
\newblock {Self-improving} reactive agents based on reinforcement learning,
  planning and teaching.
\newblock {\em Machine Learning}, 8:293--321, 1992.

\bibitem{kapturowski2018recurrent}
Steven Kapturowski, Georg Ostrovski, Will Dabney, and et~al.
\newblock {Recurrent} experience replay in distributed reinforcement learning.
\newblock In {\em International Conference on Learning Representations (ICLR)},
  2019.

\bibitem{gpu_robotic_viktor}
Jacky Liang, Viktor Makoviychuk, Ankur Handa, and et~al.
\newblock {GPU}-accelerated robotic simulation for distributed reinforcement
  learning.
\newblock {\em Proceedings of Machine Learning Research (PMLR)}, 87:270--282,
  2018.

\bibitem{mnih2013playing}
Volodymyr Mnih, Koray Kavukcuoglu, David Silver, and et~al.
\newblock {Playing} atari with deep reinforcement learning.
\newblock {\em arXiv preprint arXiv:1312.5602}, 2013.

\bibitem{dota}
OpenAI.
\newblock Openai five.
\newblock \url{https://blog.openai.com/openai-five/}, 2018.

\bibitem{alphago}
David Silver, Aja Huang, Christopher~J. Maddison, and et~al.
\newblock Mastering the game of go with deep neural networks and tree search.
\newblock {\em Nature}, 529:484--503, 2016.

\bibitem{stooke2018accelerated}
Adam Stooke and Pieter Abbeel.
\newblock {Accelerated} methods for deep reinforcement learning.
\newblock {\em arXiv preprint arXiv:1803.02811}, 2018.

\bibitem{nvas}
Oreste Villa, Daniel Lustig, Zi~Yan, and et~al.
\newblock {Need} for speed: Experiences building a trustworthy system-level gpu
  simulator.
\newblock In {\em IEEE International Symposium on High Performance Computer
  Architecture (HPCA)}, 2021.

\bibitem{vinyals2017starcraft}
Oriol Vinyals, Timo Ewalds, Sergey Bartunov, and et~al.
\newblock Starcraft ii: A new challenge for reinforcement learning.
\newblock {\em arXiv preprint arXiv:1708.04782}, 2017.

\end{thebibliography}
}
\end{document}